  \documentclass{gretsi}
 \usepackage[english,french]{babel}   
  \usepackage{times}			
  \usepackage[T1]{fontenc}
  \usepackage{amsmath}
\usepackage{amssymb}
\usepackage{multirow}
\usepackage{graphicx}
\usepackage{color}
\usepackage{ulem}
\usepackage{algorithm} 
\usepackage{algpseudocode} 
\usepackage{stmaryrd}
\usepackage{caption}
\usepackage{subcaption}

\newcommand{\addSylv}[1]{\textcolor{black}{{#1}}}
\newcommand{\addAlina}[1]{\textcolor{black}{{#1}}}

\titre{Détection de petites cibles par apprentissage profond et critère \textit{a contrario}}

\auteur{\coord{Alina}{Ciocarlan}{1},
        \coord{Sylvie}{Le Hegarat-Mascle}{2},
    \coord{Sidonie}{Lefebvre}{1},
    \coord{Clara}{Barbanson}{3}}

\adresse{\affil{1}{DOTA, ONERA, Université Paris-Saclay, F-91123 Palaiseau, France}
         \affil{2}{SATIE, Université Paris-Saclay, 91405 Orsay, France}
         \affil{3}{Safran Electronics $\&$ Defense, F-91344 Massy, France}}

\email{alina.ciocarlan@onera.fr,
sylvie.le-hegarat@universite-paris-saclay.fr\\
sidonie.lefebvre@onera.fr, clara.barbanson@safrangroup.com}

\resumefrancais{La détection de petites cibles est une problématique délicate mais essentielle dans le domaine de la défense, notamment lorsqu'il s'agit de différencier ces cibles d'un fond bruité ou texturé, ou lorsqu'elles sont de faible contraste. Pour mieux prendre en compte les informations contextuelles, nous proposons d'explorer différentes approches de segmentation par apprentissage profond, dont certaines basées sur les mécanismes d'attention. Nous proposons également d'inclure un module d'attention par canal au TransUnet, réseau à l'état de l'art, ce qui permet d'améliorer significativement les performances. Par ailleurs, le manque de données annotées induit une perte en précision lors des détections, conduisant à de nombreuses fausses alarmes non pertinentes. Nous explorons donc des méthodes \textit{a contrario} afin de sélectionner les cibles les plus significatives détectées par un réseau entraîné avec peu de données.}

\resumeanglais{Small target detection is an essential yet challenging task in defense applications, since differentiating low-contrast targets from natural textured and noisy environment remains difficult. To better take into account the contextual information, we propose to explore deep learning approaches based on attention mechanisms. Specifically, we propose a customized version of TransUnet including channel attention, which has shown a significant improvement in performance. Moreover, the lack of annotated data induces weak detection precision, leading to many false alarms. We thus explore \textit{a contrario} methods in order to select meaningful potential targets detected by a weak deep learning training.}

\begin{document}
\maketitle
\section{Introduction}
\vspace*{-.3cm}
La détection de petites cibles est un grand défi en vision par ordinateur, principalement du fait de la petite taille des cibles et de leur environnement bruité qui peut conduire à de nombreuses fausses alarmes. Quelques méthodes d'apprentissage profond ont été étudiées dans des travaux antérieurs : elles sont basées sur des réseaux de neurones convolutifs (CNN) \cite{dacremont_cnn-based_2019} et incluent parfois des mécanismes d'attention \cite{chen_local_2021}. L'un des avantages de ces derniers est qu'ils modélisent mieux les dépendances à grande échelle comparés aux CNNs. Cette propriété est un atout pour la détection de cibles, celles-ci ne présentant pas de structure spécifique. Partant de cette observation, \cite{liu_infrared_2021} utilise une version améliorée de U-Net qui inclut un encodeur Transformer (ViT) en plus de l'encodeur convolutif classique, ce qui conduit à des résultats très compétitifs. 

Une autre difficulté de cette application est le manque de données annotées pour entrainer le détecteur, ce qui résulte en une détection comportant beaucoup de fausses alarmes. En effet, le réseau de neurones n'a pas suffisamment d'exemples pour apprendre à extraire les bonnes caractéristiques. Pour autant, en observant la carte des scores donnée en sortie du réseau, les cibles y apparaissent comme étant noyées dans du bruit. Ce bruit, bien que moins significatif perceptuellement, est détecté comme cible du fait de sa forte valeur pixellique. Cela est dû au seuil fixe appliqué en sortie du détecteur pour effectuer la détection, qui ne permet pas de prendre en compte certains critères de perception comme la forme ou la densité induite par les niveaux de gris (cf. Figure \ref{fig:preds} colonne 2). Pour pallier cela, nous proposons d'explorer l'intérêt d'un critère \textit{a contrario} appliqué sur la carte des scores obtenue en sortie du détecteur afin de sélectionner les détections les plus significatives au sens du Nombre de Fausses Alarmes (NFA, défini dans le paragraphe \ref{a_contrario}). La méthode proposée permet de considérer aussi bien des caractéristiques de niveaux de gris que des éléments de structuration spatiale tels que la densité ou la forme des nuages de points représentant des cibles potentielles.

Après avoir présenté les concepts clés ainsi que l'état de l'art associé, nous décrirons une méthode de filtrage \textit{a contrario} de la carte des scores obtenue en sortie de réseaux neuronaux, dont nous analyserons l'intérêt en dernière partie.

\vspace*{-.1cm}
\section{Définitions et travaux connexes}
\vspace*{-.2cm}
\paragraph{Méthodes \textit{a contrario}}
\label{a_contrario}
Les méthodes de détection \textit{a contrario} s'inspirent des théories de la perception, en particulier celle de Gestalt \cite{desolneux2007gestalt}. Elles reposent sur le principe d'Helmoltz qui stipule qu'une grande déviation d'un modèle aléatoire est probablement dûe à la présence d'une structure. Les méthodes \textit{a contrario }consistent à rejeter un modèle naïf caractérisant un fond destructuré en choisissant un seuil de détection. Ce seuil est choisi de sorte à contrôler le Nombre de Fausses Alarmes (NFA), souvent défini en chaque \textit{objet} testé $x_i$ par :
\begin{equation}
    NFA(x_i)=N_{test}\times P(|X_i|\ge|x_i|),
\end{equation}
où $N_{test}$ représente le nombre de tests, $(X_i)_{i \in \mathbb{N}}$ une suite de variables aléatoires suivant la loi du modèle naïf et $P$ la probabilité associée. Un évènement sera alors considéré comme étant $\epsilon$-significatif si son $NFA$ est inférieur au seuil de détection $\epsilon$. 

Parmi les travaux précédents utilisant un critère NFA pour de la détection, certains travaux ont défini le modèle naïf en termes de distribution des niveaux de gris et la détection est alors réalisée en chaque pixel \cite{ipol.2019.263}, tandis que d'autres ont défini un modèle naïf en termes de distribution des pixels de valeur 1 dans une image binaire (c'est le cas de l'article ~\cite{desolneux2003grouping}).

\vspace*{-.3cm}
\paragraph{Mécanismes d'attention}
Malgré l'efficacité des CNNs pour extraire des informations significatives d'une image, l'invariance de translation induite par les convolutions semble nuire à la compréhension globale de la scène. Selon~\cite{tuli_are_2021}, cela induit un biais de texture élevé dans le processus de décision, tandis que les mécanismes d'attention semblent contourner cette limitation en imitant le cerveau et la perception humaine. 
Plusieurs types d'attention existent, conduisant à un large éventail de techniques discutées dans~\cite{guo_attention_2021}. Les méthodes les plus récentes, telles que les ViT, reposent sur des attentions spatiales. Ces dernières améliorent significativement la modélisation des dépendances entre les différentes zones d'une image, conduisant à un biais de forme élevé~\cite{tuli_are_2021}. Cependant, cela réduit la perception de petits objets, ce qui n'est pas souhaitable dans notre application. Dans cet article, nous avons décidé de nous appuyer sur des architectures hybrides CNN-Transformers à l'état de l'art, comme le TransUnet \cite{chen_transunet_2021}, afin de mieux prendre en compte l'information contextuelle tout en préservant la perception des petits objets. Parmi les autres mécanismes d'attention on peut citer l'attention par canal, qui permet de sélectionner les canaux pertinents en vue d'une réduction de la dimension. Il s'agit d'un des modules d'attention implémentés dans MA-Net \cite{fan_ma-net_2020} (bloc d'attention de fusion
multi-échelle (MFA)), en plus d'un module d'attention spatiale.
\vspace*{-.1cm}
\section{Méthodologie explorée}
\vspace*{-.1cm}
\subsection{Détecteurs de petites cibles}
Sur la base de l'état de l'art actuel, nous évaluons plusieurs approches de segmentation pour la détection de petites cibles, inspirées de l'architecture U-Net. Ces détecteurs utilisent un encodeur ResNet-18 avec en entrée une image en niveaux de gris de taille $256\times 256$ pixels. Les détecteurs sélectionnés sont :
\begin{itemize}
    \item Un U-Net, qui représente notre algorithme de référence. 
    \item Un MA-Net, dont l'architecture détaillée peut être consultée dans les Figures 1, 3 et 4 de~\cite{fan_ma-net_2020}.
    \item Un TransUnet, dont un schéma détaillé est présenté sur la Figure 1 de \cite{chen_transunet_2021}. Il s'agit d'une méthode à l'état de l'art pour la détection des cibles \cite{liu_infrared_2021}. 
    \item Un TransUnet modifié, appelé TransUnet-MFA, qui inclut le bloc MFA proposé dans le modèle MA-Net afin d'aider à la sélection des canaux pertinents lors de la concaténation au niveau du décodeur.
\end{itemize}

\vspace*{-.1cm}
\subsection{Formulation du NFA}
\vspace*{-.1cm}
\addSylv{Pour le critère NFA, nous nous sommes inspirés de~\cite{RezaeiHADM22}. L'idée principale est de considérer simultanément des caractéristiques de niveaux de gris et de structuration spatiale (densité de points). \addAlina{Nous considérons} qu'un ensemble de pixels succeptibles de représenter une cible est d'autant plus significatif qu'il contient beaucoup de points spatialement proches et de valeur élevée sur la carte des scores. Le modèle naïf est alors la distribution de Bernoulli de paramètre $p$ représentant la présence d'un pixel à une position donnée dans un espace 3D, $\mathcal{E}\subset\mathbb{R}^3$, discret et borné, d'axes représentant les coordonnées spatiales et les valeurs des scores (troisième axe) transformées (cf. paragraphe suivant). La probabilité d'observer au moins $\kappa$ pixels dans un pavé de volume $\nu$ est alors la loi Binomiale de paramètre $p$ et le NFA s'écrit :
\begin{equation}
\label{eq:nfa}
NFA\left(\kappa,\nu,p\right)=\eta \sum_{i=\kappa}^{\nu}\binom{\nu}{i} p^{i}\left(1-p\right)^{\nu-i},
\end{equation}
où $\eta$ est le nombre de tests, ici pris égal au nombre de pavés de même taille que celui considéré dans l'espace 3D. 
En termes de mise en \oe{}uvre pratique, trois éléments sont à préciser.}

\addSylv{Premièrement, les valeurs des scores sont transformées de façon à ce que les faibles valeurs soient étalées sur une forte dynamique et les scores élevés concentrés sur une faible dynamique. Dans les résultats présentés nous avons utilisé la fonction inverse : $\forall x>\tau, f(x)=\frac{1}{x-\tau}; \forall x\leq\tau, f(x)=+\infty$. En pratique le paramètre $\tau$ permet de ne pas considérer des scores trop faibles et de réduire la complexité algorithmique.}

\addSylv{Deuxièmement, plutôt que de calculer le NFA de l'Eq.~\ref{eq:nfa} qui est coûteux numériquement, nous utilisons la significativité définie par $S\left(\kappa,\nu,p\right) = - \ln(NFA_B(\kappa,\nu,p))$ et l'approximation de Hoeffding: si $\frac{\kappa}{\nu}>p$, 
\begin{equation}
\label{eq:hoef}
S\left(\kappa,\nu,p\right)\approx
\nu\left[\frac{\kappa}{\nu}\ln\left(\frac{\frac{\kappa}{\nu}}{p}\right)+\left(1-\frac{\kappa}{\nu}\right)\ln\left(\frac{1-\frac{\kappa}{\nu}}{1-p}\right)\right]-\ln\eta.
\end{equation}}


\addSylv{Troisièmement, pour calculer les nombres de points de chaque pavé de $\mathcal{E}$, nous utilisons l'histogramme intégral comme dans~\cite{Hegarat-MascleA19}. 
Finalement, l'algorithme \ref{alg:nfa} résume les principales étapes du calcul du NFA avec une astuce consistant à ne calculer la significativité que pour les pavés a priori les plus significatifs (ceux de volume minimal à nombre de points inclus donné). 
}

\begin{algorithm}[t]
	\caption{Détection de cibles de taille maximale $M$ pixels sur la carte des scores $I_s$; en entrée: $I_s$, $M$, significativité minimale $S_{min}$; en sortie: liste des cibles $\mathbf{C}$.} 
	\label{alg:nfa}
	\begin{algorithmic}[1]
		\For {pour chaque pixel $j$ de $I_s$}
			\State $I_s(j) = f(I_s(j))$
		\EndFor
		\State $\mathcal{P}\leftarrow$ nuage de points 3D issus des pixels $j \slash I_s(j)<+\infty$
        \State $p\leftarrow \frac{|\mathcal{P}|}{\text{volume 3D de } \mathcal{P}}$
        \State Initialiser un tableau $Tab$ de dimension $M$ à $+\infty$
        \State Initialiser un tableau $Idx$ de dimension $M$ à $-1$
		\For {chaque pavé 3D $\mathcal{C}$, de projection 2D (x,y) d'aire inférieure à $M$}
		   \State $\kappa\leftarrow$ nombre de pixels dans $\mathcal{C}$;  $\nu\leftarrow$ volume de $\mathcal{C}$
		   \If {$Tab[\kappa]>\nu$} 
		    \State $Tab[\kappa]\leftarrow\nu$; $Idx[\kappa]\leftarrow$ indice de $\mathcal{C}$
		    \EndIf
		\EndFor
		\For {$\kappa\in\llbracket 1,T\rrbracket$}
		    \State $\nu\leftarrow Tab[\kappa]$; $p_\mathcal{C}=\frac{\kappa}{\nu}$
            \If {$\nu<+\infty$ et $p_\mathcal{C}>p$}
                \State calculer $S(\mathcal{C})$ à partir de l'Eq.~\ref{eq:hoef}
           \EndIf
		\EndFor
		\State $\mathcal{I}\leftarrow$ liste des indices des pavés triés selon $S(\mathcal{C})$
		\State $S_{max}\leftarrow$ significativité du premier élement de $\mathcal{I}$
		\State $\mathbf{C}\leftarrow\emptyset$
		\For {chaque indice $i$ dans $\mathcal{I}$}
		    \State $\mathcal{C}_i\leftarrow$ $i^{th}$ pavé selon $\mathcal{I}$
		    \If {$S(\mathcal{C}_i)>S_{min}$ et $S(\mathcal{C}_i)>0.8\times S_{max}$}
		        \State $\mathbf{C} \leftarrow \mathbf{C}\cup\{\mathcal{C}_i\}$
		    \EndIf
		\EndFor    
	\end{algorithmic} 
\end{algorithm}

\vspace*{-.1cm}
\section{Résultats et discussions}
\vspace*{-.2cm}
En raison de la nécessité de disposer de capacités opérationnelles de jour comme de nuit, nos détecteurs de cibles ont été entraînés sur des images infrarouge annotées provenant du jeu de données d'entraînement MFIRST \cite{wang2019miss}. Deux entrainements ont été effectués : l'un en utilisant 2500 images annotées, l'autre en se plaçant dans un contexte frugal et en n'utilisant que 100 images, choisies aléatoirement. Les détecteurs sont évalués sur l'ensemble de test MFIRST ($100$ images) en termes de précision, de rappel et de score F1 calculés au niveau objet. Les résultats sont donnés en pourcentage sous la forme $\mu \pm \sigma$, avec $\mu$ la moyenne et $\sigma$ l'écart-type calculés sur cinq entrainements initialisés aléatoirement.

\vspace*{-.45cm}
\paragraph{Performance des détecteurs}

\begin{table}[t] 
\small
\centering

  \begin{tabular}{|c|c|c|c|} 
  \hline

    \ \textbf{Méthodes}    & \textbf{Prec. $\pm \sigma$(\%) } &\textbf{ Rap. $\pm \sigma$(\%) }& \textbf{F1$\pm \sigma$ (\%) }\\
    \hline
    U-Net & 50.1 $\pm$ 9.0 & 84.8  $\pm$ 2.9 & 62.5  $\pm$  7.1 \\
    MA-Net &62.1  $\pm$ 9.4 & \textbf{88.3  $\pm$ 1.2 }& 72.6  $\pm$ 6.2 \\
    TransUnet & \textbf{90.2 $\pm$ 2.7} & 81.9  $\pm$ 3.8 & 85.8  $\pm$ 1.8 \\
    TransUnet-MFA & 89.5  $\pm$ 3.1 & 86.7  $\pm$  2.1 & \textbf{88.0  $\pm$  0.6} \\
   \hline
  \end{tabular}
  \caption{Performance des détecteurs entrainés sur 2500 images, métriques calculées au niveau objet.}
  \label{res_detectors}
\end{table}

\begin{table}[t] 
\small
\begin{subtable}[h]{0.45\textwidth}
\centering

\begin{tabular}{|p{1.0cm}|c||c|c|c|} 
    
  \hline

      \multicolumn{2}{|c||}{\textbf{Méthodes}}  & \textbf{Prec. $\pm \sigma$ (\%)} & \textbf{Rap. $\pm \sigma$(\%)}
     & \textbf{F1 $\pm \sigma$(\%)}\\
    \hline
    \hline
     \multirow{3}{*}{U-Net}& S & 20.7  $\pm$ 13.9 & \textbf{53.6  $\pm$ 33.5} & 29.7  $\pm$ 19.2 \\ 
     & S+F & 35.6$\pm$28.3 & 43.4$\pm$39.2 &35.4$\pm$33.2 \\
      & NFA & \textbf{76.9$\pm$14.3} & 37.7$\pm$26.1 & \textbf{44.9$\pm$23.5} \\
     \hline
     
    \multirow{3}{*}{MA-Net} & S & 45.7  $\pm$ 11.6 & \textbf{80.3  $\pm$2.5} & 57.4  $\pm$ 9.5 \\ 
     & S+F &  47.0$\pm$12.0 & 79.6$\pm$1.4 & 58.3$\pm$9.7 \\
      & NFA & \textbf{74.6$\pm$4.8} & 68.1$\pm$3.1 & \textbf{71.2$\pm$2.8} \\
      \hline
   
  \end{tabular}
   \caption{Entrainement sur 100 images.}
       \label{tab:train100}
  \end{subtable}
  \hfill
  \begin{subtable}[h]{0.45\textwidth}
  \centering
  
\begin{tabular}{|p{1.3cm}|c||c|c|c|} 
  \hline

      \multicolumn{2}{|c||}{\textbf{Méthodes}}  & \textbf{Prec. $\pm \sigma$(\%)} & \textbf{Rap.$\pm \sigma$ (\%)}
     & \textbf{F1 $\pm \sigma$(\%)}\\
    \hline
    \hline
     
     \multirow{3}{*}{U-Net}& S & 50.1 $\pm$ 9.0 & \textbf{84.8  $\pm$ 2.9} & 62.5  $\pm$  7.1\\ 
     & S+F &  79.2$\pm$7.8 &83.6$\pm$2.5 & 81.1$\pm$3.7 \\
      & NFA &  \textbf{84.9$\pm$6.6} & 81.9$\pm$4.2 & \textbf{83.1$\pm$2.0} \\
     \hline
    \multirow{3}{*}{TransUnet-} & S  & 89.5  $\pm$ 3.1 & 86.7  $\pm$  2.1 & 88.0  $\pm$  0.6 \\ 
      & S+F &  90.7$\pm$2.7 & 86.4$\pm$1.9 & \textbf{88.5$\pm$0.7} \\
     MFA & NFA & \textbf{93.4$\pm$1.8} & 78.3$\pm$3.0 & 85.1$\pm$2.1 \\
   \hline
  \end{tabular}
   \caption{Entrainement sur 2500 images.}
       \label{tab:train2500}
  \end{subtable}
  
  \caption{Performance des détecteurs  entrainés sur 100 ou 2500 images. Les prédictions sont données soit par seuillage fixe (S), par filtrage du seuillage fixe (S+F) ou par calcul du NFA. Métriques calculées au niveau objet.}
  \label{tab:res_detectors_frugal}
  
\end{table}

\begin{figure}[t]

    \includegraphics[width=8.5cm]{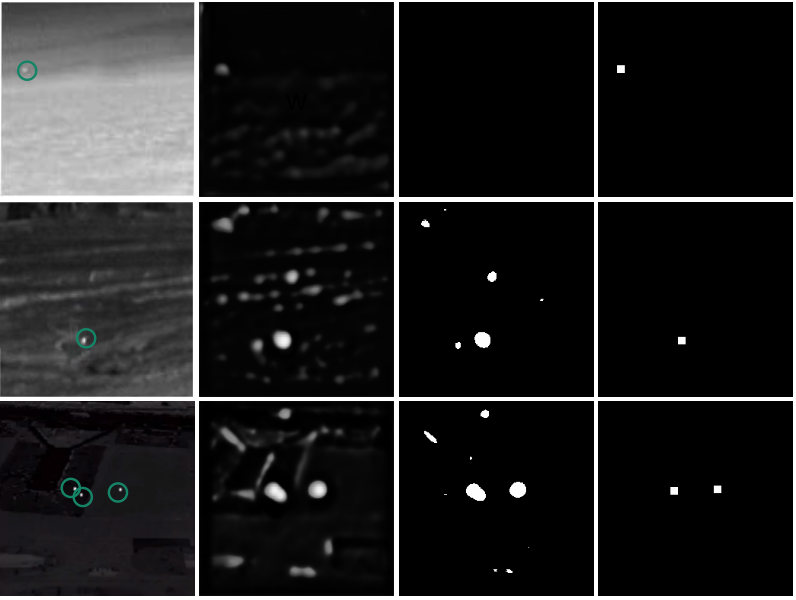}
    \caption{Exemples de prédictions des détecteurs sur 3 images (lignes). De gauche à droite : image d'origine avec la vérité terrain encerclée, carte des scores en sortie du réseau, prédiction après seuillage avec seuil fixe, prédiction après NFA.}
    \label{fig:preds}
\end{figure}
Le Tableau~\ref{res_detectors} donne les métriques obtenues lors de l'évaluation des détecteurs de cibles entrainés sur un nombre suffisant d'images (2500). Il y a un écart notable en précision lorsque l'on compare le U-Net d'origine aux versions modifiées incluant des mécanismes d'attention. Ces dernières conduisent également à de meilleures performances globales, en particulier pour TransUnet-MFA qui obtient un score F1 moyen de $88.0\%$. On remarque également que les mécanismes d'attention, en particulier ceux spatiaux, réduisent fortement la variabilité des résultats. Le Tableau~\ref{tab:train100} donne les résultats lorsque l'on se place dans un contexte frugal pour U-Net et MA-Net (100 images d'entrainement). 
Dans ce cas, on remarque que la précision est très faible pour le cas où l'on applique un seuil fixe (S) aux cartes de score en sortie des réseaux de neurones pour détecter les cibles, ce qui conduit à un taux de fausses alarmes élevé. La variabilité des résultats est également très élevée notamment du fait d'un cas de non convergence parmi les cinq expériences pour le U-Net. La Figure \ref{fig:preds} permet de visualiser quelques résultats pour trois images. La première colonne montre l'image d'origine, la suivante la carte de sortie des réseaux, la troisième le résultat de prédiction en appliquant un seuil fixe et la dernière en y appliquant un critère NFA. On y voit les nombreux faux positifs induits par le bruit de fond présent dans la carte des scores, bien qu'ils soient moins pertinents perceptuellement que les cibles à détecter.
\vspace*{-.5cm}
\paragraph{Contribution du NFA}

Pour pallier ce problème, nous appliquons un calcul de NFA sur les cartes de score pour effectuer les détections, et nous comparons les résultats avec ceux obtenus par un filtrage morphologique (ouverture) de la carte de détection donnée après seuillage par un seuil fixe (S+F). Nous avons testé les detecteurs U-Net et MA-Net dans un contexte frugal du fait de leur très mauvaise précision dans ce contexte. Le Tableau~\ref{tab:res_detectors_frugal} donne les résultats obtenus pour les différentes méthodes testées. Nous avons aussi évalué deux détecteurs entraînés sur 2500 images et ayant une meilleure précision.

Nous pouvons observer que, dans un contexte frugal, le NFA améliore fortement la précision, ce qui réduit le taux de fausses alarmes. Cela donne lieu à de meilleures performances globales comparé au seuillage fixe (S) ou à son filtrage (S+F). La Figure~\ref{fig:preds} rend compte de ces améliorations.

En revanche, lorsque le détecteur est suffisamment performant pour donner en sortie des cartes de scores avec peu de bruit, l'apport du NFA est plus limité par rapport à un filtrage avec un seuil fixe\addSylv{:} d'après le Tableau \ref{tab:train2500} la précision est augmentée de manière significative, au détriment du nombre de bonnes détections. \addSylv{En effet, par définition le NFA controlant le nombre de fausses alarmes va optimiser la précision des résultats (ce qui est bien observé sur le Tableau~\ref{tab:res_detectors_frugal}, tant au niveau de la précision moyenne que de sa variabilité) et non le F1 score.} Par ailleurs, cette limitation provient également d'une extraction d'information réalisée par le détecteur de manière erronée. La cible ne ressortant pas sur la carte des scores, le calcul de NFA ne pourra pas réparer l'erreur commise en amont. Ainsi, le NFA aura plus d'intérêt pour les sorties de réseaux où la cible se trouve noyée dans un bruit, ce qui est le cas pour les réseaux faiblement entrainés. Dans des travaux futurs, nous envisageons d'intégrer un critère \textit{a contrario} à un réseau de neurones dès l'apprentissage afin que les avantages du NFA puissent être bénéfiques à un plus grand nombre de scénarios.
\vspace*{-.2cm}
\section{Conclusion}
\vspace*{-.2cm}
Nous avons exploré plusieurs approches par apprentissage profond pour la détection de petites cibles, basées notamment sur les mécanismes d'attention. Celles-ci donnent des résultats globaux très compétitifs, mais manquent de précision dans un contexte frugal. Pour contrer ce problème, nous avons remplacé le seuillage fixe effectué sur les cartes de scores par un critère \textit{a contrario}. Les conclusions de ces travaux préliminaires constituent une motivation pour explorer l'intégration d'un critère NFA dans la fonction de coût des réseaux.
\vspace*{-.2cm}
\bibliographystyle{ieeetr}

\bibliography{biblio}

\end{document}